\documentclass[runningheads]{llncs}
\usepackage{graphicx}
\usepackage{color}
\usepackage{multirow}
\usepackage{authblk}
\usepackage{amsmath,amssymb} 
\usepackage{floatrow}
\usepackage{breqn}
\usepackage{subcaption}
\begin{document}

\title{Large Scale Scene Text Verification with Guided Attention} 
\titlerunning{Large Scale Scene Text Verification with Guided Attention} 

\author{Dafang He\textsuperscript{1}\footnote[1]{The two authors contribute equally}, Yeqing Li\textsuperscript{2}$^*$, Alexander Gorban\textsuperscript{2}, Derrall Heath\textsuperscript{2}, Julian Ibarz\textsuperscript{2}, Qian Yu\textsuperscript{2}, Daniel Kifer\textsuperscript{1}, C. Lee Giles\textsuperscript{1}}    

\institute{The Pennsylvania State University\textsuperscript{1}, Google Inc\textsuperscript{2}.\\
	\email{ \{duh188,giles\}@ist.psu.edu}, \email{dkifer@cse.psu.edu}, \email{ \{yeqing,dheath,gorban,julianibarz,qyu\}@google.com}
}

\authorrunning{Large Scale Scene Text Verification with Guided Attention}

\maketitle

\begin{abstract}
Many tasks are related to determining if a particular text string exists in an image.
In this work, we propose a new framework that learns this task in an end-to-end way. 
The framework takes an image and a text string as input and then outputs the probability of the text string being present in the image.
This is the first end-to-end framework that learns such relationships between text and images in scene text area.
The framework does not require explicit scene text detection or recognition and thus no bounding box annotations are needed.   
It is also the first work in scene text area that tackles such a weakly labeled problem.
Based on this framework, we developed a model called \textit{Guided Attention}.
Our designed model achieves better results than several state-of-the-art scene text reading based solutions for a challenging Street View Business Matching task.
The task tries to find correct business names for storefront images and the dataset we collected for it is substantially larger, and more challenging than existing scene text dataset.
This new real-world task provides a new perspective for studying scene text related problems.
\end{abstract}

\section{Introduction}
\vspace{-2pt}
\footnotetext{The work is done while Dafang is in Internship at Google.}
Our streets are full of text such as street names, street numbers, store names, etc.
These text are important clues for understanding the real world.
With this rapidly growing new source of data, it has become more and more important to learn how to extract useful text information out of images.
This task is usually referred to as scene text reading.

\begin{figure}
\floatbox[{\capbeside\thisfloatsetup{capbesideposition={right,top},capbesidewidth=7cm}}]{figure}[\FBwidth]
{\caption{An example application of scene text verification task. Our model takes a business storefront image and a potential business name, and then directly outputs how likely the storefront image matches the business name. The red text means it is the ground truth for this image and our model gives it a high score.}\label{fig:intro_image}}
{\includegraphics[width=5cm]{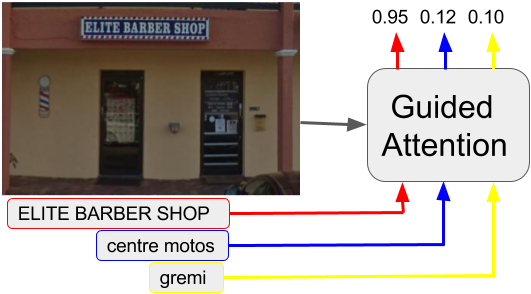}}
\end{figure}

Many researchers divide the scene text reading problem into two sub-problems: text detection~\cite{chen2004detecting,epshtein2010detecting,neumann2012real,zhang2016multi,He_2017_CVPR,Shi_2017_CVPR} and text recognition~\cite{AAAI1612256,jaderberg2016reading,shi2016robust}.
In order to build such an applicable system, the two components are usually combined sequentially into a complex system even though they are developed separately.
In many applications, we don't need to precisely transcribe every single word in an image - we only need to tell if some text prominently exists in an image and we call this problem: \textbf{scene text verification}.
If we use scene text reading to formulate it,  we will be facing several challenges:
(1) Building a robust text understanding system is difficult, especially considering the possible differences in the domains of images in real usage and in training.
A detector trained on currently available public datasets (e.g. ICDAR 2015~\cite{karatzas2015icdar}) may have high possibility of failure on the new domain of images the system is deployed on.
(2) Scene text is normally highly artistic. Heuristic based text matching is usually applied afterwards for verification, and often error are introduced when connecting these pieces.
(3) A large amount of fully annotated data is needed in order to train a detector, which is expensive to obtain.
In this work, we argue that we do not need separate text detection and text recognition systems for text verification problem.

We have observed that, in a lot of situations, an explicit text transcriber is unnecessary.
Often a context is available, which can be used to narrow down the attention (text candidates).
For example, if a person wants to find the reviews of a restaurant, they may want to just take a picture of the restaurant and we should be able to provide the reviews for them.
In such cases, the algorithm only needs to pay attention to the name of the restaurant in order to identify it.
It does not need to read any other text on the street, such as a 20\% off advertising on clothes, or the opening hours on the door.
Furthermore, a list of restaurant candidates could be obtained based on the geo-location so as to serve as candidates.
Therefore, the actual problem we are trying to solve is instead: how likely is it that a sequence of words is in the image?
In answering this question, an undiscriminating transcriber may be harmful since it could provide extra transcriptions that are irrelevant to the question (i.e., noise), and/or provide incorrect transcriptions which confuse the later process.

In order to address the previous concerns, we propose a new, end-to-end framework which takes input the image and the text and predict the possibility that the text string is contained in an image as in Fig.~\ref{fig:intro_image}.
The framework is expected to give a high probability if the input sequence is in the image, and thus it tries to verify the existence of a text string.
It could also be regarded as trying to find a match between an image and a text string.
Many applications could be built when we are able to give such a unified text-image score.
For example, In addition to the restaurant case, Google Street View contains countless images taken on the street.
Such a model could enable it to identity the business name from storefront images for back-end data processing.

We design a model called \textit{Guided Attention} based on such framework.
It does not need to do any explicit text detection or recognition.
Instead it uses an attention mechanism that is guided by the input text string and decides whether the string is in the image or not.

This is the first work that tries to solve this problem in an end-to-end manner, and we study it in the context of business matching---given a storefront image, predict which business it represents.
We collect millions of business storefront images through Google Maps API~\footnote[1]{https://developers.google.com/maps/} for this task.
Each image is associated with a list of candidate business names and the algorithm needs to find correct candidate among them.

We call this the Street View Business Matching (SVBM) dataset, and experiments show the effectiveness of our new framework in solving this challenging real-world problem. Our contributions are:

\begin{enumerate} 
\item \textbf{New Problem} We study a new problem: scene text verification in an image.
It verifies the existence of a certain text string and is closely related to scene text reading.
It could be tackled with traditional method based on scene text detection and scene text recognition.
It could also been seen as a sub-problem for scene text retrieval.

\item \textbf{New Dataset} We collect a large-scale dataset(SVBM) which contains 1.5M images for experiment.
It is substantially larger than any existing public scene text datasets(e.g ICDAR2015 contains only 1500 images).
The dataset contains storefront images from google street view, and algorithm needs to verify the existence of the business name in the images.
It contains various of different languages and is much more challenging than existing public scene text datasets.

\item \textbf{New Framework} We propose an end-to-end trainable framework to this problem that does not do or require any explicit text detection or recognition.
The framework is completely different from traditional method, and only requires image-text pairs for training, as opposed to a fully annotated dataset (e.g., bounding box labels), which are more expensive to create than simple yes/no labels.
This greatly reduces the burden of dataset collection.
We also propose a model based on this new framework and it achieves better performance in terms of precision/recall curve than existing text reading based solutions on SVBM.
This new framework brings new insight into scene text area, and could inspire further research works.

\item \textbf{Extensive Experiments} We have done extensive experiments to study different properties of the framework.
Our major experiments are on SVBM dataset, and we also evaluate the trained model in two public datasets: UberText~\cite{UberText} and FSNS~\cite{smith2016end}.
These experiments confirm that our Guided Attention model is better suited at solving this task than a more traditional scene text reading based solution and by combing the proposed method and traditional method, we can achieve an even better performance in this task.
We have also done ablation studies that show how important some design aspects of our model impact performance.
\end{enumerate}

\section{Related Work}
Scene text detection is usually considered the most challenging part of building an end-to-end system, and many techniques have been developed to solve it.
There are usually two approaches. 
The first approach treats the problem as a semantic segmentation problem, and several further steps are performed to obtain each individual text instance.
This category includes methods that use fully convolutional neural networks (FCNs) to do semantic segmentation on text to extract text blocks~\cite{zhang2016multi,He_2017_CVPR,Zhou_2017_CVPR}.
The second category uses modified region proposal networks (RPNs) to detect text~\cite{tian2016detecting,Shi_2017_CVPR}.
Both approaches achieve state-of-the-art performance on public datasets.

The challenges of scene text recognition are mainly due to the fact that text could be distorted in an image.
Most regular (straight) text could be effectively read by first extracting CNN features and then using a RNN to predict the sequence.
~\cite{AAAI1612256,shi2016end} used such techniques and combined them with CTC loss~\cite{graves2006connectionist} to create powerful models for transcribing straight text.
~\cite{wojna2017attention,shi2016robust,ijcai2017-458} tried to read curved scene text by using an attention mechanism, and spatial attention~\cite{wojna2017attention,ijcai2017-458} has achieved the state-of-the-art performance.

Scene text retrieval~\cite{mishra2013image,karaoglu2017words}, which aims at retrieving images based on text content, is closely related to scene text verification.
The verification task could be seen as a subtask for scene text retrieval, as it only cares about the existence of text and no ranking is needed.
Retrieval performance will benefits from a better verification model.

An end-to-end scene text reading system would combine the detection and recognition steps sequentially, and a fully annotated dataset is needed to train it.
Instead, our proposed framework takes as input a character string and an image, and then verifies the existence of the string in the image in an end-to-end manner. 
Our framework is unique in that no explicit text detection or recognition is involved, and thus, there is also no need for a fully annotated dataset.
The annotation for our framework is thus much easier to obtain since we only need images with corresponding lists of candidate strings and their labels.

Our work also has a loose connection to visual question answering\cite{antol2015vqa} or image text matching\cite{yan2015deep}.
However, our framework also has several unique properties:
(1) input text is character sequence instead of words.
(2) Simple binary classification is used as evaluation metric.
(3) Order of input words shouldn't affect the results. For example, if a sequence of words ``Street View Image'' is in an image, then ``Image Street View'' should also be considered as positive.

In experiments, we evaluate our framework and model on the SVBM dataset and two public datasets: UbetText~\cite{UberText}, FSNS~\cite{smith2016end}.
SVBM dataset contains image-text pairs without bounding box annotations and it makes training a scene text detector impossible.
The model has to learn from both the image and text in order to give a final prediction.

\section{Method}
\subsection{Model Architecture}

\begin{figure*}
\begin{center}
 \includegraphics[width=0.9\textwidth,height=5cm]{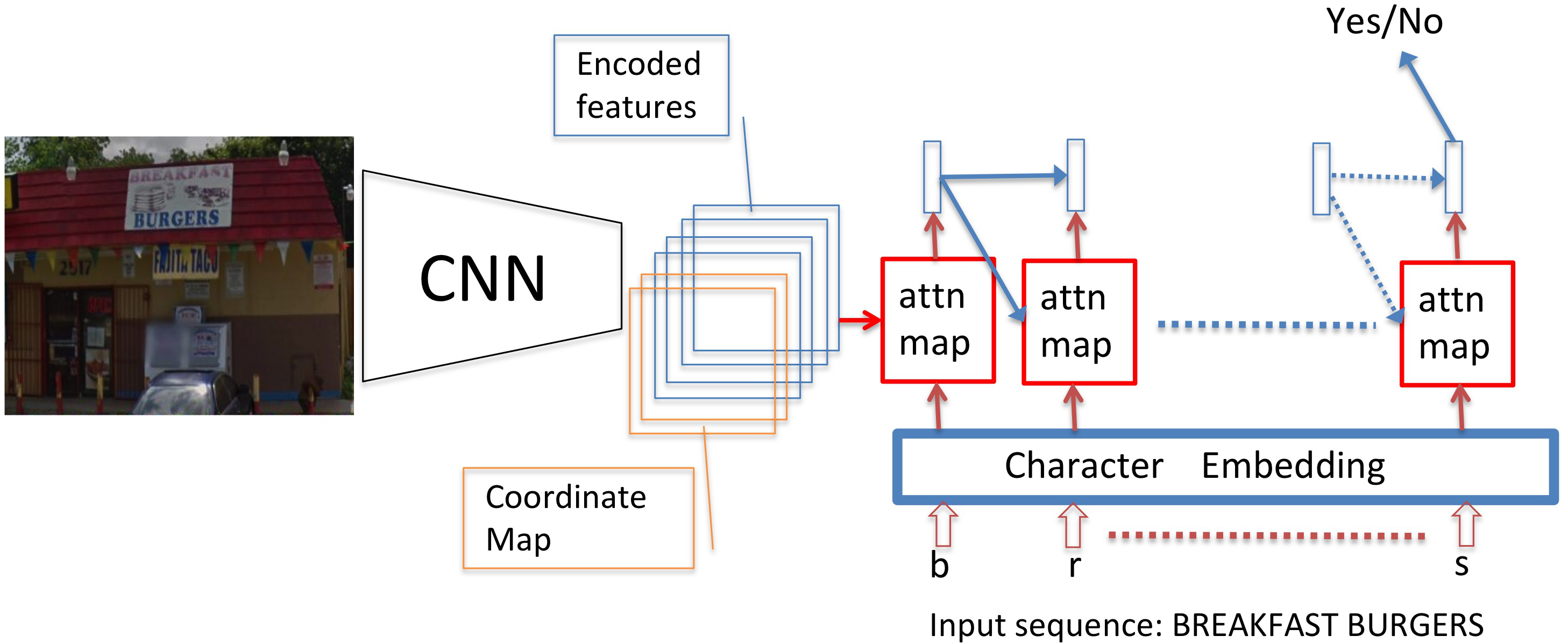}
\end{center}
   \caption{The architecture of our scene text verification model. The model takes in a sequence of characters and an image, and predict the probability of the input text sequence being present in the input image. }
\label{fig:Architecture}
\end{figure*}

The architecture is shown in Fig.~\ref{fig:Architecture}. 
Our model consists of two major components: (1) a CNN-based image encoder network with coordinate encoding, (2) a guided attention decoder which selectively pools features from the encoded feature map and generates the final result.

\vspace{-4pt}
\subsubsection{CNN Encoder with Coordinate Map}
We trimmed InceptionV3 \cite{szegedy2016rethinking} to construct our image encoder, which builds a deep CNN by stacking some carefully designed sub-network structure with different scales.

\begin{figure}
\floatbox[{\capbeside\thisfloatsetup{capbesideposition={right,top},capbesidewidth=5cm}}]{figure}[\FBwidth]
{\caption{Illustration of encoding the coordinates as a one-hot feature vector $f_{xy}$ and concatenate it with the original visual features $f_v$. A feature map with H $\times$ W $\times$ C will have corresponding coordinate feature map with size H $\times$ W $\times$ (H+W)}\label{fig:coordinate_encoding}}
{\includegraphics[width=6cm]{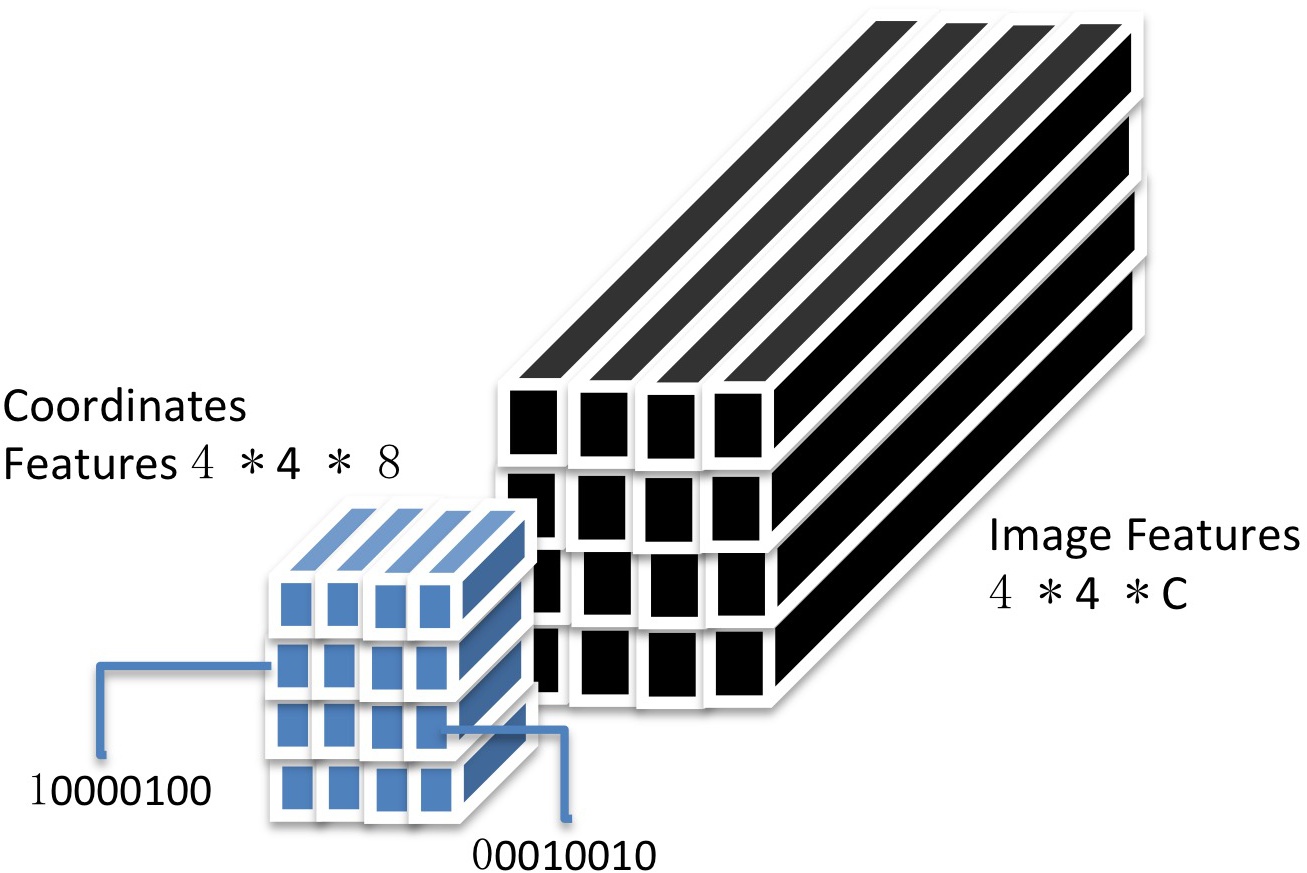}}
\end{figure}

Let $I$ be the input image, and CNN encoded visual feature is denoted as $f_v=\text{CNN}(I)$.
In order to capture the spatial information for each location in the feature map, we follow~\cite{wojna2017attention} to create a coordinate map $f_{xy}$ for each image.
The equation of such coordinate encoding could be expressed as $f_{xy} = Eocode(i, j)$.
$i,j$ denotes the x, and y indices of each spatial location in the feature map.
Function $Encode$ computes the one-hot encoding for each positing $i,j$, and we concatenate the coordinate map $f_{xy}$ with original cnn feature $f_v$ in depth dimension.
We use $\tilde{f}$ to denote the features augmented with position information.
By combining such position information with each feature vector, the following decoding process could refer to them for better attention calculation and achieves better decoding results.
Such scheme has been adopted by~\cite{wojna2017attention,ijcai2017-458} and has been proved to be effective in scene text related tasks.
Fig~\ref{fig:coordinate_encoding} illustrates the coordinate encoding.

\vspace{-4pt}
\subsubsection{Guided Attention Decoder}
In the next step, the model tries to decode useful information from the cnn features $\tilde{f}$ guided by the input character string.
Let $N$ be the number of characters of the input string and $S={S_1, \ldots, S_N}$ be the character level one-hot encoding of it.
The embeddings of the characters are represented as $S^e={S^e_1, \ldots, S^e_N}$, which is learned end-to-end during training.
Our goal is to compute
\begin{align}
p_{valid} = \mathbb{P}(y=1|S_1, \ldots, S_N, I),
\label{eq:pvalid}
\end{align} 
where $y \in \{0, 1\}$ is the indicator of the existence of the text.

We use LSTM~\cite{hochreiter1997long} as our recurrent function to encode sequential features.
Let us denote $h_t$ as the hidden state of the LSTM in time step $t$, then the update function of hidden state could be expressed as Eq. \ref{eq:update_h}:
\begin{align}
	h_{t} = \mathrm{LSTM}(h_{t-1}, S^e_t, \mathrm{CT}(x_t)),
\label{eq:update_h}
\end{align} 
where $\mathrm{CT}(x_t)$ represents the context vector generated in time step $t$.
It can be computed based on Eq.~\ref{eq:Select_pool}:
\begin{align}
	\mathrm{CT}(x_t) = \sum\limits_{i=1}^W \sum\limits_{j=1}^H \alpha_{(i,j)}^t * \tilde{f}_{(i,j)},
\label{eq:Select_pool}
\end{align} 

where $\alpha_{(i,j)}^t$ represents the attention map.
It is computed based on Eq.~\ref{eq:Align_Fn}.
$e^t_{(i,j)}$ represents how relevant is the feature $\tilde{f}_{(i, j)}$ to the current character embedding $S^e_t$.
In this work, we choose to use attention function as in \cite{vinyals2015grammar} in Eq.~\ref{eq:Attn_Fn}:
\begin{align}
	\alpha_{(i,j)}^t = \frac{\exp(e^t_{(i,j)})}{\sum\limits_{u=1}^W \sum\limits_{v=1}^H \exp(e^t_{(u,v)})}, \nonumber \\
	e^t_{(i,j)} = f_{attn}(h_{t-1}, \tilde{f}_{(i, j)}) \label{eq:Align_Fn}, \\
f_{attn}(h_{t-1}, \tilde{f}_{(i, j)}) = v^T \tanh(Wh_{t-1} + U\tilde{f}_{(i, j)}) \label{eq:Attn_Fn},
\end{align} 
where $W$, $V$ are weight matrix that could be learned.

Let us denotes $y={y_1, \ldots, y_N}$ as the output sequence probability and $y'={y_1', \ldots, y_N'}$ as the groundtruth labels.
Each $y_i$ is the prediction based on information till character $S^e_i$.
In training, we use cross entropy as our loss function, and we only calculate the loss based on the output of the last time step for each image and candidate pair as in Eq.~\ref{eq:Loss_fn}: 
\begin{align}
 loss = - \frac{1}{M} \sum\limits_{i=1}^M (y_i' \log(y^i_{n_i}) + (1-y_i') \log((1-y^i_{n_i}))),
\label{eq:Loss_fn}
\end{align} 
where $n_i$ is the length of the $i$th candidate and M is the number of training pairs.

\subsection{Model Training and Sub-batch Architecture}
In our problem setting, each image usually contains several positive (random shuffling of positive words) and a list of negative text.

In order to save the computation, for each image we have $M$ parallel text input for it and the CNN tower will only need to be computed once.
The number $M$ is equal to $N_n + N_p$ where $N_p$, $N_n$ represent the number of positive and negative examples sampled for each image, respectively.
The total loss thus becomes a weighted cross entropy based on the ratio between positive examples and negative examples. 
In our experiment, we usually set $N_p$ to be 1, and $N_n$ to be 4.
So there are 5 parallel recurrent sequences for each convolutional tower.

\vspace{-2pt}
\subsection{Hard Negative Mining(HNM)}
\begin{figure}
\floatbox[{\capbeside\thisfloatsetup{capbesideposition={right,top},capbesidewidth=5.5cm}}]{figure}[\FBwidth]
{\caption{Example showing how we can determine the hardness of a negative example based the corresponding positive candidates. The different color represents the different hardness levels.}\label{fig:HardNegExample}}
{\includegraphics[width=6cm]{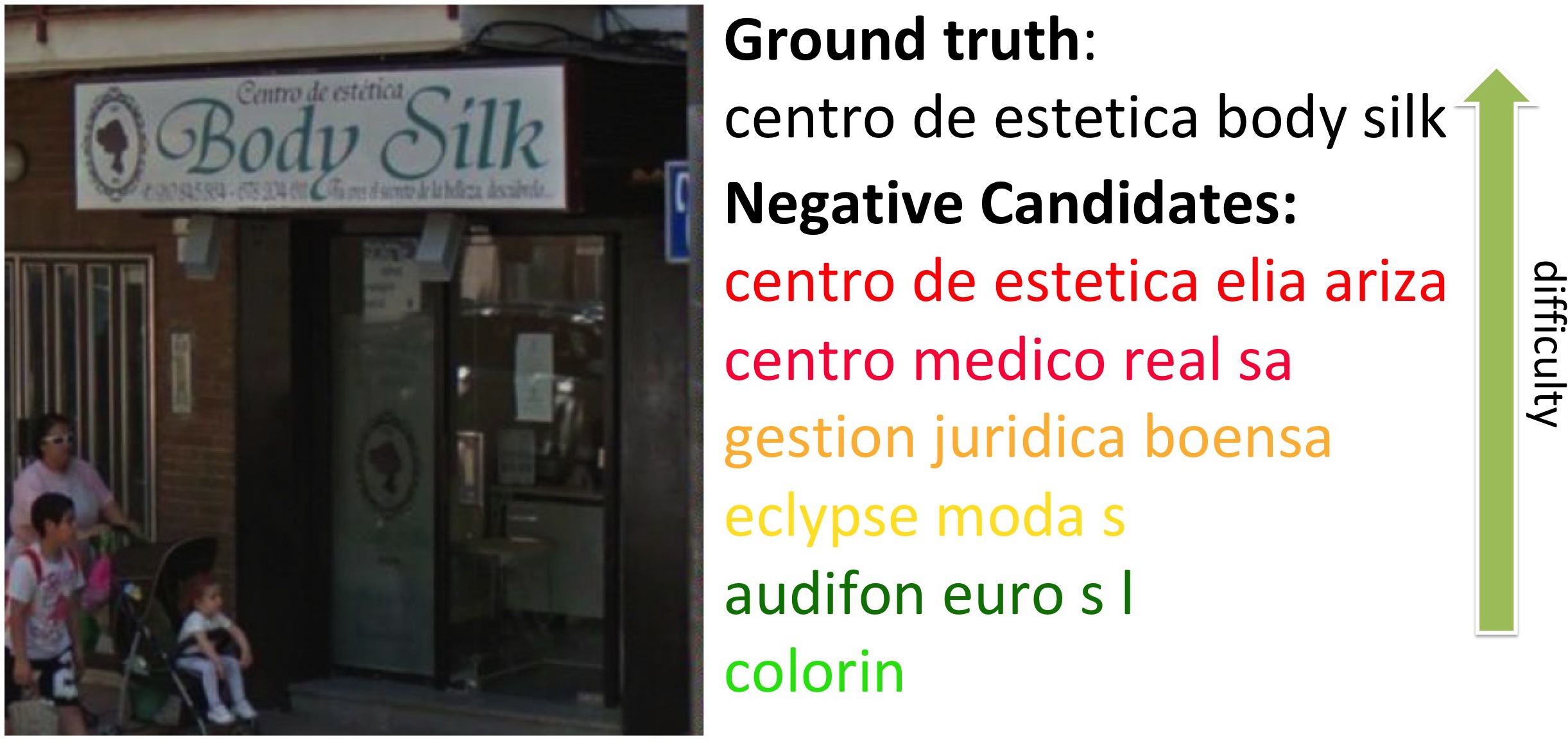}}
\end{figure}

\vspace{-3pt}
During model training, the sampling scheme of the $N_n$ number of negative training examples plays an important role for better performance.
In our problem, the hardness of a negative candidate could be empirically determined by the edit distance of it with the corresponding positive candidates.
Fig. \ref{fig:HardNegExample} illustrates this idea.

We also observe that, in the SVBM dataset, most negative candidates are easy cases as they differ from the positive text by quite a lot.
However, the negative samples that really confuse the network are from those hard cases.
So we incorporate hard negative example mining as a fine tuning process for our model.
We first define the hardness of an example based on Eq.~\ref{eq:Hardness}.

\begin{dmath}
  \mathrm{Hardness}(\hat{neg}) = 1 - \min_{\forall \hat{pos} \in \widehat{P}} \frac{\mathrm{edit\_dis}(\hat{neg}, \hat{pos})}{\max(\mathrm{len}(\hat{pos}), \mathrm{len}(\hat{neg}))}
\label{eq:Hardness}
\end{dmath}

$\widehat{P}$ represents the positive example set for a specific image.
$\hat{pos}$ is one positive sample and $\hat{neg}$ represents the negative sample that we want to compute.
The function $\mathrm{edit\_dis}$ calculates the edit distance between the positive sample $\hat{pos}$ and the negative sample $\hat{neg}$.
The hardness of a negative sample is determined by the minimal edit distance between the negative candidate and all the positive samples.
The higher the score, the harder the negative sample.

The training process including HNM is thus as follows:
(1) Train the network from scratch with evenly sampled positive and negative text for each image.
(2) Finetune the network with evenly sampled positive text and evenly sampled negative text whose hardness score is larger than $T$.
In our experiments, we set $T = 0.3$ to keep relatively harder text without removing too much of them.
Note that we did this for both training set and testing set, so in the second phase, the testing examples are harder.
During the first phase, the classification accuracy reached $95\%$, but in the beginning of the second phase, the classification accuracy dropped to $88\%$. 
This means that the harder examples, based on our definition, are actually more difficult for our model.

\section{Experiments}
The major experiments are on the SVBM dataset which contains millions of images, and we use it to study different properties of our model as well as the problem of scene text verification.
We have also done two experiments on UberText and FSNS datasets with the model trained on SVBM dataset and baseline methods.

\subsection{Dataset Description and Experiment Setting}
\subsubsection{SVBM} 
The SVBM dataset is based on Street View images, and each image represents one business storefront.
The storefront images could be obtained by the method proposed in~\cite{yu2015large}, and
all the images have associated geo-locations (for e.g, lat/lng) so that we can retrieve a list of candidate business names by filtering out far-away business.
The number of candidates depends on the business density and the search radius.
The goal is to find the correct business name among the list of candidates.
No text bounding boxes are provided.

The dataset contains 1.39M training images and around 147K testing images.
They are collected from various countries such as US, Brazil, England and etc. with various different languages.
Each image has a number of candidates ranging from 10 to over 500.
One of them is the correct business name, and all others are incorrect.
As a preprocessing step, we convert all business names into lower case.
So the character set contains a total of 37 alphanumeric characters (including space) plus characters in other languages which have high frequency in training set.
We evaluate the dataset based on precision recall curve on a per-candidate basis.

In training, we use rmsprop as our optimization method, and the learning rate is set to 0.001.
The batch size is set to 32, and the input image is resized and padded to 400 * 400 with random noise for simplicity.
Each image is associated with 5 candidate text during training, and thus the batch size for attention decoder is 160(32 * 5).
We use 70 cpu machines for training and it takes over 20 days to converge to an initial checkpoint.

\subsubsection{UberText and FSNS}
There is currently no public dataset that is suitable for our purpose.
We choose two public datasets that are relatively larger with a little bit different evaluation schemes.
This study aims at demonstrating that by combining the proposed model with other traditional approaches, a better performance can be achieved for the text verification problem. 

\textit{UberText}~\cite{UberText} is a newly released dataset for scene text related problems.
It is currently the largest fully annotated public dataset and contains 10K testing images. 
We evaluate it using the model trained on SVBM to show that our model can generalize well to other datasets.
We choose the words that are of type: ``business names" in the dataset and only evaluate the recall of these positive text in the verification problem.

\textit{FSNS}~\cite{smith2016end} contains french street signs.
The images are much easier than that in SVBM because the text are focused and clear.
We randomly sample 49 text as negative text for each image for evaluation purpose.

\subsection{Qualitative Results} 
\begin{figure*}[th]
\definecolor{DarkGreen}{rgb}{0,0,0}
\definecolor{DarkRed}{rgb}{0.4,0,0}
 \tiny 
    \center
    \scalebox{1.0}{
    \setlength\tabcolsep{2pt}
    \begin{tabular} {p{0.6in} p{0.2in} p{0.6in} p{0.2in} p{0.6in} p{0.2in} p{0.6in} p{0.2in} p{0.6in} p{0.2in}} 
\multicolumn{2}{c}{
      \includegraphics[width=0.18\linewidth]{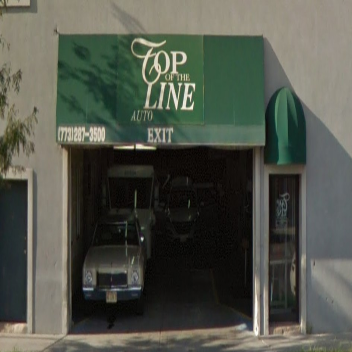}
      }
   & \multicolumn{2}{c}{
      \includegraphics[width=0.18\linewidth]{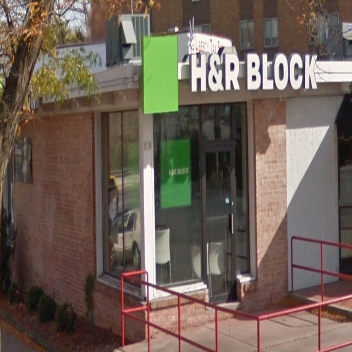}
      }
   & \multicolumn{2}{c}{
      \includegraphics[width=0.18\linewidth]{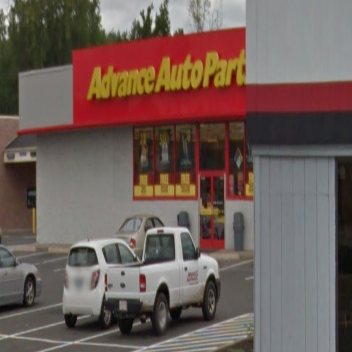}
      }
   & \multicolumn{2}{c}{
      \includegraphics[width=0.18\linewidth]{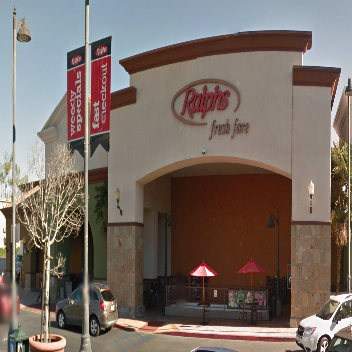}
      }
   & \multicolumn{2}{c}{
      \includegraphics[width=0.18\linewidth]{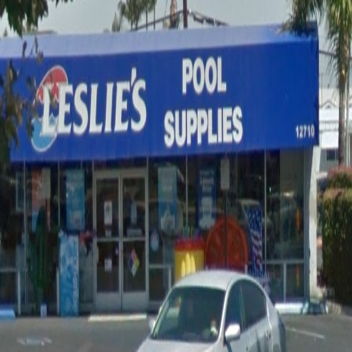}
      }
   \\
*Top of the .. & 0.696 &*H\&R Block & 0.734 &*Advance .. & 0.637 & *Ralphs & 0.921 & *Leslie's ... & 0.945 \\
Pcrr & 0.287 & Us Bank & 0.280 & Access to .. & 0.213 & KC Produ... & 0.338 & Hungry n .. & 0.567\\
Ed's Hot .. & 0.237 & CW Price & 0.193 & ATM (Webs ... & 0.176 & Cactusbird .. & 0.202&  A1 Auto .. & 0.437 \\
Christian .. & 0.158 & ~ & ~ & ~ & ~ &Studio Plaza & 0.132 & ~ & ~ \\
R\&B Conco.. & 0.142 & ~ & ~ & ~ & ~ &Work Boot .. & 0.124 & ~ & ~ \\
Candelas .. & 0.140 & ~ & ~ & ~ & ~ & Cardtronics .. & 0.081 & ~ & ~ \\
\multicolumn{2}{c}{
      \includegraphics[width=0.18\linewidth]{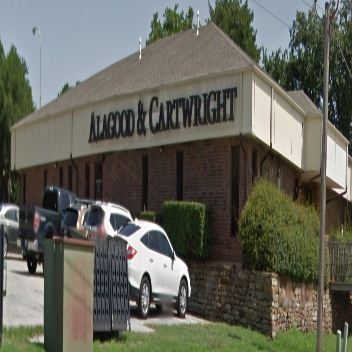}
      }
   & \multicolumn{2}{c}{
      \includegraphics[width=0.18\linewidth]{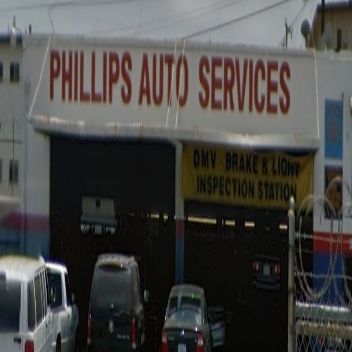}
      }
   & \multicolumn{2}{c}{
      \includegraphics[width=0.18\linewidth]{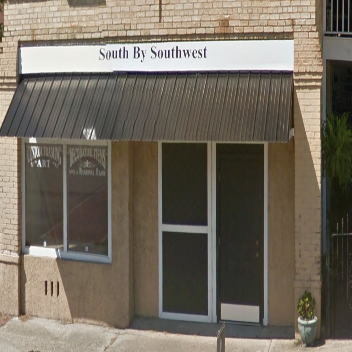}
      }
   & \multicolumn{2}{c}{
      \includegraphics[width=0.18\linewidth]{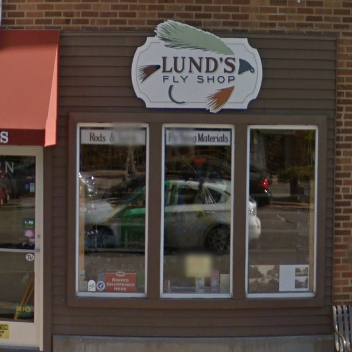}
      }
   & \multicolumn{2}{c}{
      \includegraphics[width=0.18\linewidth]{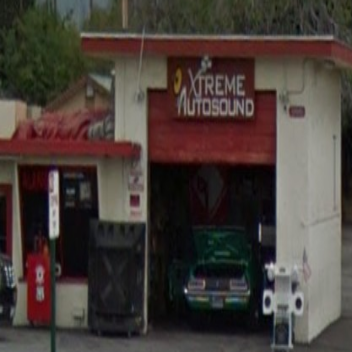}
      }
   \\
*Alagood .. & 0.804 & *Phillips .. & 0.962  & *South by .. & 0.945 & *Lund's Fly .. & 0.862 & *Xtreme Au.. & 0.693 \\
Taylor Sar .. & 0.276 & Dry Down .. & 0.246& Teapot Mus.. & 0.201 & The Marti .. & 0.443& Rock The .. & 0.271 \\
Denton Cou .. & 0.155 & Coupon & 0.240 & Old Town .. & 0.148 & ATM (BP) & 0.366 &  ATM & 0.073 \\
~ & ~ & Fresh \& Cl.. & 0.151 & Scherer Da .. & 0.086 & Lazy River & 0.338 & ~ & ~ \\
~ & ~ & Plaza Segundo & 0.150& ~ & ~ &First Bapti.. & 0.230& Just ATMs Inc & 0.223 \\
~ & ~ & The San Remo & 0.146& ~& ~ &River Falls & 0.167 & ~ & ~ \\
\multicolumn{2}{c}{
      \includegraphics[width=0.18\linewidth]{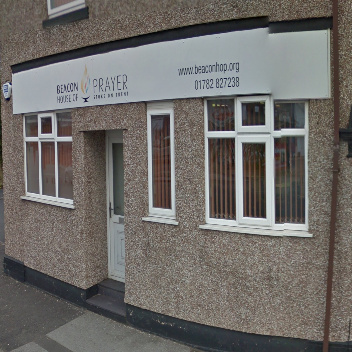}
      }
   & \multicolumn{2}{c}{
      \includegraphics[width=0.18\linewidth]{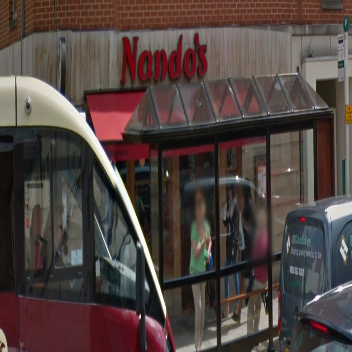}
      }
   & \multicolumn{2}{c}{
      \includegraphics[width=0.18\linewidth]{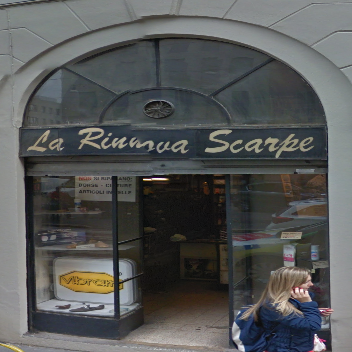}
      }
   & \multicolumn{2}{c}{
      \includegraphics[width=0.18\linewidth]{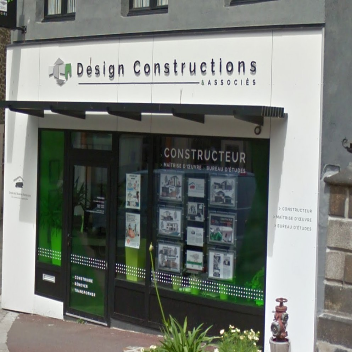}
      }
   & \multicolumn{2}{c}{
      \includegraphics[width=0.18\linewidth]{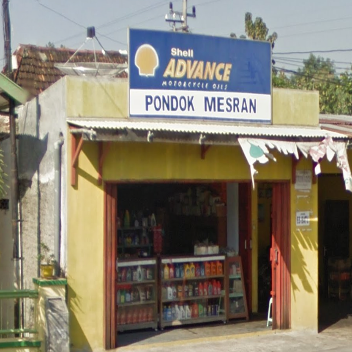}
      }
   \\
*beacon hou .. & 0.722 &*Nando's .. & 0.935& *La Rinno .. & 0.963&*Design Con .. & 0.957&*Pondok Mes.. & 0.941\\
Select & 0.155&The Child .. & 0.179& Civiche racc... & 0.249&Tabac - Pres... & 0.295 &Beauty Spa 2 & 0.581\\
Sinar Kaca & 0.575&Scope & 0.141& Borse.Pro .. & 0.162&Matrix Fe .. & 0.270 & *T-Mobile & 0.119 \\
~ & ~ &CEX - Ent ... & 0.135 &Intesa Sa .. & 0.154 &Edf Gdf & 0.270&Toko Hous .. & 0.333\\
~ & ~ &Station Way & 0.118& DE NIGRIS .. & 0.141 &Tri Slectif & 0.202&Simbah. Net & 0.210\\
~ & ~ &Vodafone & 0.109&Il Aria Folli & 0.141&Studio Coiff... & 0.166&SD Negeri Si... & 0.206  \\
\multicolumn{2}{c}{
      \includegraphics[width=0.18\linewidth]{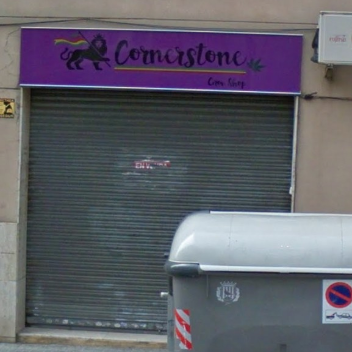}
      }
   & \multicolumn{2}{c}{
      \includegraphics[width=0.18\linewidth]{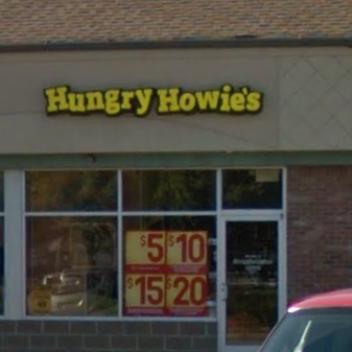}
      }
   & \multicolumn{2}{c}{
      \includegraphics[width=0.18\linewidth]{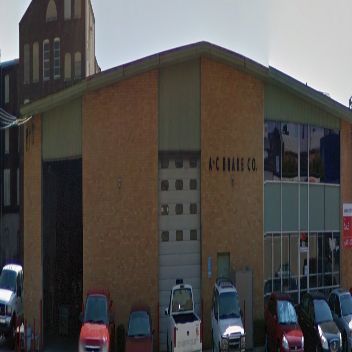}
      }
   & \multicolumn{2}{c}{
      \includegraphics[width=0.18\linewidth]{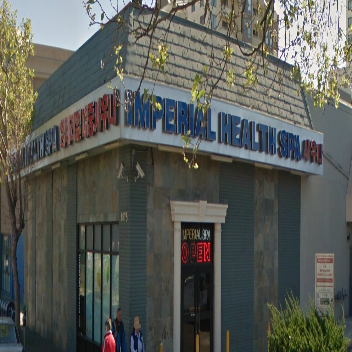}
      }
   & \multicolumn{2}{c}{
      \includegraphics[width=0.18\linewidth]{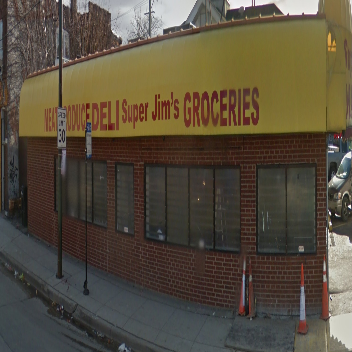}
      }
   \\
Avila & 0.684&*Hungry Ho .. & 0.419&*A-C Brake .. & 0.507& The People .. & 0.627 & New Kingdom .. & 0.219 \\
Riko's & 0.219&Efmark Dep .. & 0.193&Christ Wa .. & 0.481& Holy Dog & 0.580 & ATM Express & 0.189 \\
Tradicionarius & 0.203&Cardtronics .. & 0.076&~&~& Gene Suttle .. & 0.389 & *Super Gia.. & 0.170 \\
*Cornerst .. & 0.201& ~ & ~ & ~ &~ & Queen's Hou .. & 0.369 & ~ & ~ \\
Jordi's & 0.177& ~ & ~ & ~ & ~ & Jane & 0.350 & ~ & ~ \\
A Cantina .. & 0.177& ~ & ~ & ~ & ~ & Nare & 0.305 & ~ & ~ 
    \end{tabular}
    }
    \vspace{-0.05in}
    \caption{Example images from the test set. We only visualize the top predicted candidates. The first three rows show the examples that we give good predictions(high score for positive candidate, low score for negative ones). The last row shows several failure examples. They are either caused by wrong prediction from our model, or by the problem of extremely unclear or non-exists text. It also shows that the dataset is challenging as it contains many practical problems such as extreme distortion. Such problems make it hard to correctly detect and transcribe the text. Our model, without the need to explicitly detect text, can achieve a better performance.}
    \label{fig:visual_examples}
\end{figure*}

In this section, we give several visual results illustrating the different properties of our trained guided attention model.
More visual examples are in Fig.~\ref{fig:visual_examples}.

\begin{figure*}[ht!]
\scriptsize
   \begin{tabular}{c|p{0.9in}p{0.35in}|p{0.9in}p{0.35in}|p{0.5in}p{0.1in}}
    \multirow{4}{*}{
	\includegraphics[height=0.15\linewidth,width=0.15\linewidth]{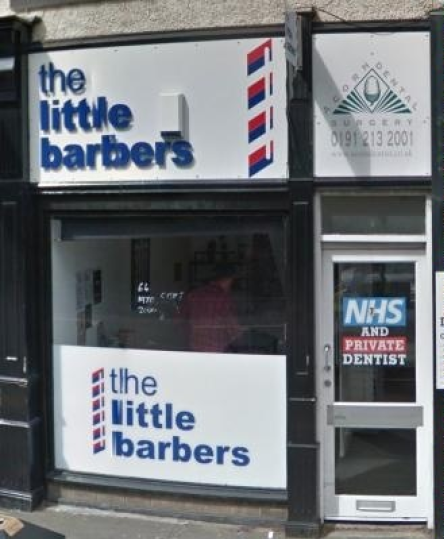} 
	}& 
The Little Barbers&0.9625 & The Little Barbers & 0.9625 &NHS & 0.8947
\\
& Little Barbers &0.9524 & Little Barber The & 0.8749 &PRIVATE
& \multirow{2}{*}{0.07}
\\
& The &0.4268 & Barber Little The & 0.9249 & DENTIST &
\\
& Little &0.4317 &Barber The Little& 0.9517& google&0.2026
\\
& Barber &0.9624 &Little The Barber &0.9614 & &
\\
& Barber shop&0.1306 &  & & &
   \end{tabular}

   \caption{A qualitative evaluation on several specific behaviors of our model when we give it the same input image, but different input text. On the left is the input image. The first section in the table on the right shows results when we input a subset of ground truth text. The 2nd section in the table demonstrates when we have randomly shuffled groundtruth text as input. The last section represents results when the input text are not the corresponding business name, but are also in the image.}
   \label{fig:QualitativeExample}
\end{figure*}

\subsubsection{Subset of Text}
During training and subsequent evaluations, we use the full business name.
However, it is interesting and important to study the property of our model when we use subsets of the business name, or a slightly different business name to see how it performs.
This is because that during annotation, we may not obtain exactly the same text as the actual business name.
For example, ``The Little Barbers" might be annotated as ``Little Barbers", and our model is expected to still give a high score for that.
The first section in the table of Fig.~\ref{fig:QualitativeExample} shows an example image and several such input text strings with their predicted probabilities.

We can see that, when we use informative subset text (For e.g, ``Little Barbers") from the ground truth as input text to the model, the model still gives a pretty high score. 
This meets our expectation as we care about the existence of the string (words) in the image.
However, there are also several interesting findings:
(1) If we use non-informative words as input(e.g: ``The"), the model gives a relatively low score.
(2) If we use text string that contains other words that are not in the image (For e.g, ``Barber Shop"), the model gives a low score.
These findings are interesting, and we believe that the model is looking for more informative words and makes decisions based on that. 
``The" is common in business names, so the model gives a lower score if we only use that as input text.
``Barber shop" contains other words, which could possibly indicate that it's some other business' name.
So our model gives a low score to it even though it contains the word ``Barber'' that is in the image.

\subsubsection{Random Shuffled Text}
As we have discussed before, our model should be able to ignore the shuffling of input words.
The 2nd section of the table in Fig.~\ref{fig:QualitativeExample} shows the results when we try to input shuffled text with the same image into the model.
This property is also important, as the order of words in annotation has no relationship with the spacial ordering of those words in image, and thus the trained model should ignore the word order, and only focus on the existence of the collection of words.

We can see that our model is somewhat invariant to the random shuffling of words as all the of them received high scores.
This is the property that we expect, and it is important to text verification.

\subsubsection{Non-Business Text}
The last section of the table in Fig.~\ref{fig:QualitativeExample} shows an example where we have text inputs that are not the specific business name in the ground truth.
We can see that the model still gives it a high score since it is in the image, and it also meets our expectation.
There is also a failure example when we have ``PRIVATE DENTIST'' as input.
This might happen when text are too small w.r.t the image size, so the attention could not capture it well.
In addition to that, since our model is trained on business related text, we believe that this also caused the failure of this non-business text.

\renewcommand{\arraystretch}{0.9}
\begin{figure}[ht!]
\footnotesize
   \begin{tabular}{c|p{1.2in}p{0.35in}|p{1.2in}p{0.35in}}
    \multirow{6}{*}{
	\includegraphics[height=0.2\linewidth, width=0.2\linewidth]{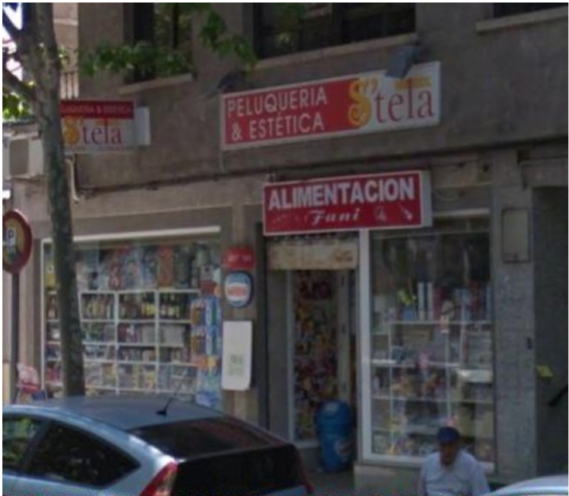}
	}&\multicolumn{2}{c}{\footnotesize \textcolor{blue}{Before Hard Negative Mining}}&\multicolumn{2}{c}{\footnotesize \textcolor{blue}{After Hard Negative Mining}}\\
&\textcolor{red}{{\scriptsize  Peluqueria estetica stela}}&{\scriptsize 0.9324}&\textcolor{red}{{\scriptsize Peluqueria estetica stela}}&{\scriptsize 0.8924}\\
&{\scriptsize peluqueria toni}&{\scriptsize 0.7613}&{\scriptsize peluqueria toni}&{\scriptsize 0.3293}\\
&{\scriptsize francisco hidalgo tello}&{\scriptsize 0.2643}&{\scriptsize francisco hidalgo tello}&{\scriptsize 0.1375}\\
&{\scriptsize ferreteria mheva sl} &{\scriptsize 0.2232}&{\scriptsize ferreteria mheva sl} &{\scriptsize 0.1328}\\
&{\scriptsize puertodental sl}&{\scriptsize 0.1989}&{\scriptsize puertodental sl}&{\scriptsize 0.1276}\\
\hline
\multirow{6}{*}{
	\includegraphics[height=0.2\linewidth, width=0.2\linewidth]{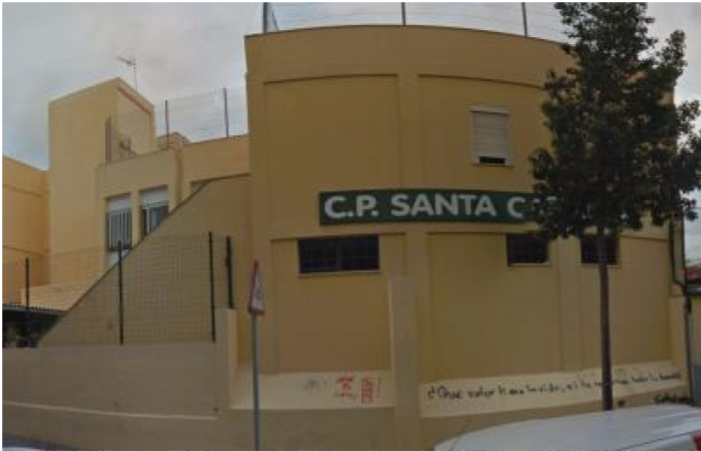}
	}&\multicolumn{2}{c}{\footnotesize \textcolor{blue}{Before Hard Negative Mining}}&\multicolumn{2}{c}{\footnotesize \textcolor{blue}{After Hard Negative Mining}}\\
	
&\textcolor{red}{{\scriptsize ceip santa catalina}}&{\scriptsize 0.9324}&\textcolor{red}{{\scriptsize ceip santa catalina}}&{\scriptsize 0.8604}\\
&{\scriptsize aulario santa catalina}&{\scriptsize 0.9057}&{\scriptsize aulario santa catalina}&{\scriptsize 0.1294}\\
&{\scriptsize trastero 16}&{\scriptsize 0.0781}&{\scriptsize verde limo}&{\scriptsize 0.1186}\\
&{\scriptsize ropa africana} &{\scriptsize 0.0618}&{\scriptsize la pizarra} &{\scriptsize 0.0970}\\
&{\scriptsize verde limon} &{\scriptsize 0.0606}&{\scriptsize trastero 16} &{\scriptsize 0.0868}\\
&{\scriptsize peluqueria vanitas} &{\scriptsize 0.0605}&{\scriptsize apartamento primas} &{\scriptsize 0.0819}\\
   \end{tabular}
   \vspace{0.1cm}
   \caption{Two visual examples illustrating the performance gain after mining hard negative candidates for training. We observe that after HNM, the gap between the best prediction and the second best prediction has usually been increased.}
   \label{fig:HardNegCompare}
   \vspace{-0.16in}
\end{figure}

\begin{figure*}[htp]
\centering
  \subcaptionbox{\label{fig:pr_compare_transcription}}{\includegraphics[width=1.5in, height=1.4in]{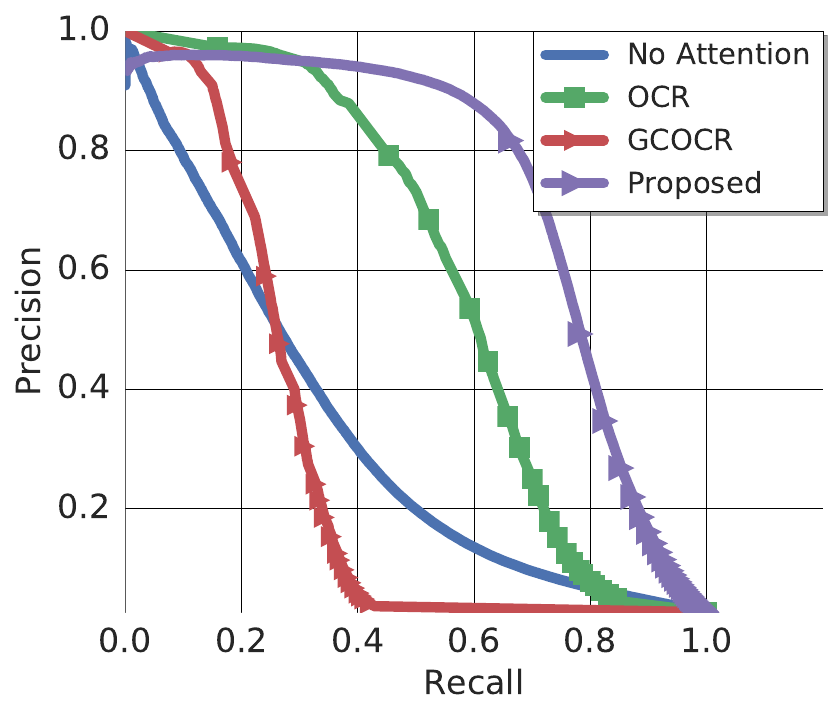}}~
  \subcaptionbox{\label{fig:pr_cut_off_len}}{\includegraphics[width=1.5in,height=1.4in]{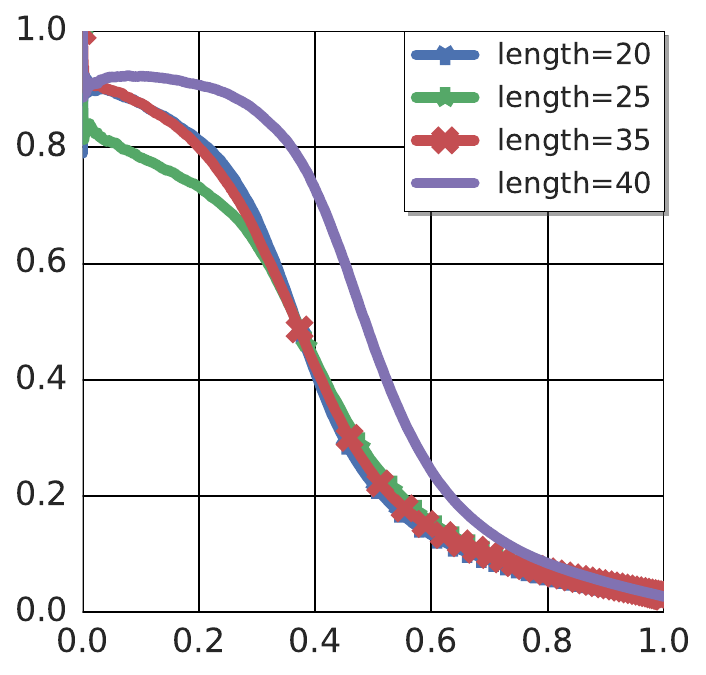}}\hspace{1em}~
  \subcaptionbox{\label{fig:pr_hard_neg_mine}}{\includegraphics[width=1.5in,height=1.4in]{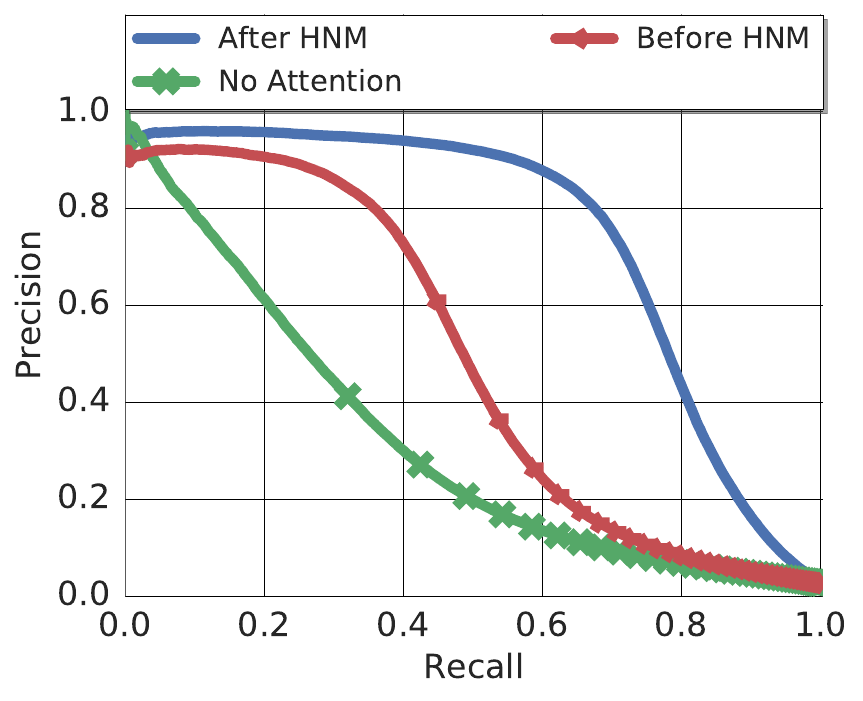}}
  
  \caption{(a) The Precision-Recall curve of our model compared with other baselines. (b) The Precision-Recall curve of our model w.r.t. different maximum length of text. (c) The Precision-Recall curve comparison before and after HNM.}
\end{figure*} 

\subsubsection{Hard Negative Mining}
Fig. \ref{fig:HardNegCompare} shows two examples illustrating the performance of the model trained before and after hard negative mining.
We can see that after hard negative mining, the probability margins between the positive and best negative sample for the two images have increased by $40\%$.
This makes the model trained with HNM much more robust and the quantitative results of such comparison are in the followings.

\subsection{SVBM Quantitative Evaluation}
\subsubsection{Baseline Models}
We compare several baseline models with our approach: (1) Google Cloud OCR(GCOCR)~\footnote[1]{https://cloud.google.com/vision/docs/ocr}. (2) Attention OCR(OCR)~\cite{wojna2017attention}. (3) Show and Tell Model~\cite{vinyals2015show} with binary classification output.
See supplementary material for details of the third model.

Model (1) tries to detect and read all the text indiscriminately.
Then we do text based matching with the candidates to find the best candidate.
This is one typical paradigm for this problem, but the model is trained on other fully annotated dataset.
Model (2) is trained on the SVBM dataset directly.
We simply force it to transcribe the whole image into the positive business name it represents.
Text based matching is performed afterwards.
Model (3) is a modification based on~\cite{vinyals2015grammar} (by changing the output as a binary classification).
It could also be regarded as removing the attention mechanism in our framework.
So we call it ``no attention".

\subsubsection{Comparison w.r.t baselines}
We first show the comparison of our final model against other baseline methods.
Fig. \ref{fig:pr_compare_transcription} shows the Precision/Recall curve.
We can see that our end-to-end model(after HNM) outperforms all other baselines by a large margin.
This address our two points:
(1) text detector and recognizer trained on other fully annotated data couldn't achieve good results in our SVBM dataset because they couldn't accurately find and transcribe the business name.
(2) Our task could be learned in an end-to-end way, and it outperforms transcription based method(OCR).
In the following two evaluations, we show different settings that can improve the performance.

\subsubsection{Evaluation w.r.t Maximum Length}
The maximum length of the character sequence could be tuned as an hyperparameter. 
This is an interesting aspect of our scene text verification problem.
It is because when we need to decide which candidate business name the storefront represents, sometimes we only need, for example, the first 20 characters.
In the example Fig~\ref{fig:HardNegExample}, we only need the first 15 characters to determine which candidate is the positive one.

Note that this maximum length $N$ only affects those candidates with length longer than it.
We simply cut the longer text to keep the first $N$ characters.
This is also a difference between our problem w.r.t traditional VQA problem, since usually we will not cut the question in VQA task.

The Precision/Recall curve w.r.t the maximum length is in Fig. \ref{fig:pr_cut_off_len}.
We can see that the peak is when the maximum length set to 40 for our task. 
However, other lengths also achieve reasonable performance.
Besides, the value is dataset dependant.
This experiment aims at illustrating an unique property of our problem and is important in deployment since the shorter the sequence, the faster the model could run.
So there is a trade-off between the performance and efficiency. 
We use 40 as our final model choice.

\subsubsection{Evaluation w.r.t Hard Negative Mining}
Whether to use hard negative mining makes a big difference in terms of the performance of the trained model.
In Fig. \ref{fig:pr_hard_neg_mine} we show the results of the models trained before and after hard negative mining.
Together we also show the results produced by a model with ``no attention''.
We can see that our attention mechanism is important to get good performance for our scene text verification tasks.
By further adding hard negative mining process, the performance has been further improved, especially in the high accuracy region of the curve.
This is also a difference between our problem with standard VQA problem since we can empirically define the ``hardness''.

\begin{figure*}[htp] 
\centering 
  \subcaptionbox{\label{fig:UberTextResults}}{\includegraphics[width=1.5in, height=1.4in]{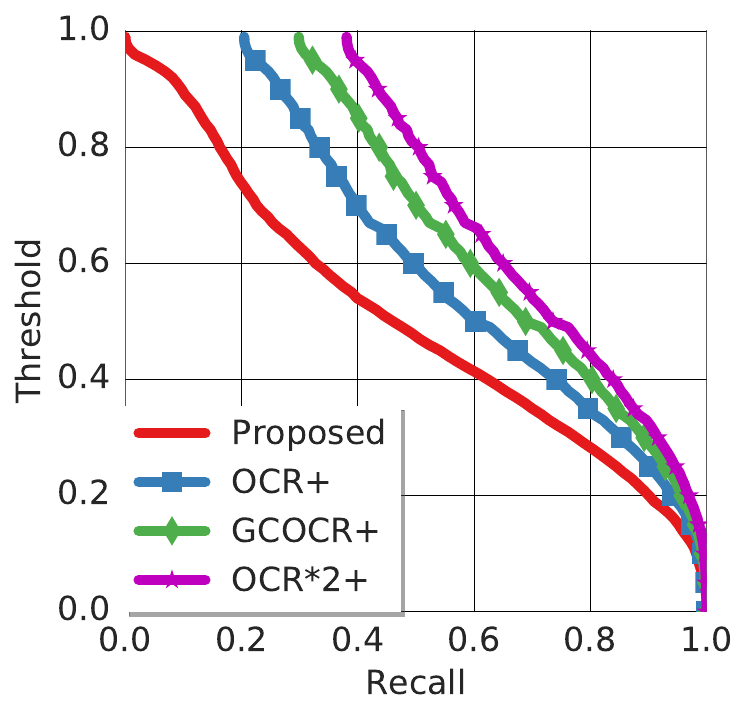}}~
  \subcaptionbox{\label{fig:FSNSResults}}{\includegraphics[width=1.5in,height=1.4in]{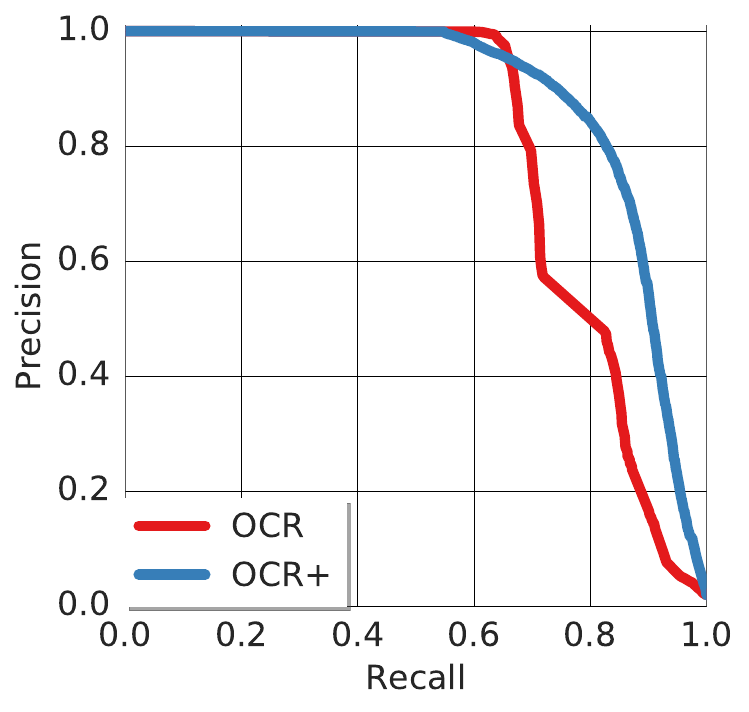}}\hspace{1em}   
  \caption{(a) The recall w.r.t threshold for different models on UberText. We only evaluate positive text here. (b) P-R curve for OCR~\cite{wojna2017attention} and the ensembled model on FSNS.}
  \vspace{-0.15in}
\end{figure*}

\subsection{UberText and FSNS Quantitative Evaluation}
\vspace{-5pt}
In this quantitative study, we study the problem that whether we can combine this model with traditional approaches to better solve text verification problem.
The compared ensembled approaches are following:
(1) OCR+: Ensemble of Attention OCR~\cite{wojna2017attention} and the proposed approach by taking the maximum of the scores output by the two models as the final score. This way of ensmeble is used in all of the following compared approaches.
(2) GCOCR+: Simple ensemble of GCOCR\footnotemark[1] and the proposed approach.
(3) OCR*2+: Simple ensemble of GCOCR\footnotemark[1], OCR~\cite{wojna2017attention} and the proposed approach.

Fig. \ref{fig:UberTextResults}, \ref{fig:FSNSResults} show the results for UberText and FSNS, respectively.
The results show that when combining our model with other traditional approaches, we can achieve a better results than the original model itself.
It also demonstrates that the ``knowledge'' learned from different models are not the same, and our model could also serve as complementary resources of information for traditional approaches.

\vspace{-5pt}
\section{Conclusion and Acknowledgement}
\vspace{-5pt}
In this work, we proposed a new problem: verifying the existence of a text string in an image and collected a large-scale dataset(SVBM) for evaluating the performance on this problem.
Instead of traditional approaches based on scene text reading, we propose an end-to-end framework which takes both the image and the text as inputs and gives a unified result.
We experimentally proved that model designed based on this framework achieved a better performance in SVBM for matching the business with the storefront image.
It could also be combined with traditional methods into more powerful models as shown in the experiments in two public datasets.
This work does not aim at giving the most sophisticated architecture, but proving that an end-to-end solution could be developed for such task to achieve a better performance without the need to fully annotate images at the text level.

This work was majorly done while Dafang is at internship at Google. We also thank NSF grant CCF 1317560 for support.

\renewcommand{\arraystretch}{0.9}

\clearpage
\bibliographystyle{splncs04}
\bibliography{egbib}

\begin{thebibliography}{10}
\providecommand{\url}[1]{\texttt{#1}}
\providecommand{\urlprefix}{URL }
\providecommand{\doi}[1]{https://doi.org/#1}

\bibitem{antol2015vqa}
Antol, S., Agrawal, A., Lu, J., Mitchell, M., Batra, D., Lawrence~Zitnick, C.,
  Parikh, D.: Vqa: Visual question answering. In: Proceedings of the IEEE
  international conference on computer vision. pp. 2425--2433 (2015)

\bibitem{chen2004detecting}
Chen, X., Yuille, A.L.: Detecting and reading text in natural scenes. In:
  Computer Vision and Pattern Recognition,CVPR. Proceedings of the IEEE
  Computer Society Conference on. vol.~2, pp. II--366. IEEE (2004)

\bibitem{epshtein2010detecting}
Epshtein, B., Ofek, E., Wexler, Y.: Detecting text in natural scenes with
  stroke width transform. In: Computer Vision and Pattern Recognition,CVPR.
  Proceedings of the IEEE Computer Society Conference on. pp. 2963--2970. IEEE
  (2010)

\bibitem{graves2006connectionist}
Graves, A., Fern{\'a}ndez, S., Gomez, F., Schmidhuber, J.: Connectionist
  temporal classification: labelling unsegmented sequence data with recurrent
  neural networks. In: Proceedings of the 23rd international conference on
  Machine learning. pp. 369--376. ACM (2006)

\bibitem{He_2017_CVPR}
He, D., Yang, X., Liang, C., Zhou, Z., Ororbi, II, A.G., Kifer, D., Lee~Giles,
  C.: Multi-scale fcn with cascaded instance aware segmentation for arbitrary
  oriented word spotting in the wild. In: The IEEE Conference on Computer
  Vision and Pattern Recognition (CVPR) (July 2017)

\bibitem{AAAI1612256}
He, P., Huang, W., Qiao, Y., Loy, C., Tang, X.: Reading scene text in deep
  convolutional sequences. In: AAAI Conference on Artificial Intelligence
  (2016)

\bibitem{hochreiter1997long}
Hochreiter, S., Schmidhuber, J.: Long short-term memory. Neural computation
  \textbf{9}(8),  1735--1780 (1997)

\bibitem{jaderberg2016reading}
Jaderberg, M., Simonyan, K., Vedaldi, A., Zisserman, A.: Reading text in the
  wild with convolutional neural networks. International Journal of Computer
  Vision  \textbf{116}(1),  1--20 (2016)

\bibitem{karaoglu2017words}
Karaoglu, S., Tao, R., Gevers, T., Smeulders, A.W.: Words matter: Scene text
  for image classification and retrieval. IEEE Transactions on Multimedia
  \textbf{19}(5),  1063--1076 (2017)

\bibitem{karatzas2015icdar}
Karatzas, D., Gomez-Bigorda, L., Nicolaou, A., Ghosh, S., Bagdanov, A.,
  Iwamura, M., Matas, J., Neumann, L., Chandrasekhar, V.R., Lu, S., et~al.:
  Icdar 2015 competition on robust reading. In: Document Analysis and
  Recognition (ICDAR), 2015 13th International Conference on. pp. 1156--1160.
  IEEE (2015)

\bibitem{mishra2013image}
Mishra, A., Alahari, K., Jawahar, C.: Image retrieval using textual cues. In:
  Proceedings of the IEEE International Conference on Computer Vision. pp.
  3040--3047 (2013)

\bibitem{neumann2012real}
Neumann, L., Matas, J.: Real-time scene text localization and recognition. In:
  Computer Vision and Pattern Recognition,CVPR. Proceedings of the IEEE
  Computer Society Conference on. pp. 3538--3545. IEEE (2012)

\bibitem{Shi_2017_CVPR}
Shi, B., Bai, X., Belongie, S.: Detecting oriented text in natural images by
  linking segments. In: The IEEE Conference on Computer Vision and Pattern
  Recognition (CVPR) (July 2017)

\bibitem{shi2016end}
Shi, B., Bai, X., Yao, C.: An end-to-end trainable neural network for
  image-based sequence recognition and its application to scene text
  recognition. IEEE transactions on pattern analysis and machine intelligence
  (2016)

\bibitem{shi2016robust}
Shi, B., Wang, X., Lyu, P., Yao, C., Bai, X.: Robust scene text recognition
  with automatic rectification. In: Proceedings of the IEEE Conference on
  Computer Vision and Pattern Recognition. pp. 4168--4176 (2016)

\bibitem{smith2016end}
Smith, R., Gu, C., Lee, D.S., Hu, H., Unnikrishnan, R., Ibarz, J., Arnoud, S.,
  Lin, S.: End-to-end interpretation of the french street name signs dataset.
  In: European Conference on Computer Vision. pp. 411--426. Springer (2016)

\bibitem{szegedy2016rethinking}
Szegedy, C., Vanhoucke, V., Ioffe, S., Shlens, J., Wojna, Z.: Rethinking the
  inception architecture for computer vision. In: Proceedings of the IEEE
  Conference on Computer Vision and Pattern Recognition. pp. 2818--2826 (2016)

\bibitem{tian2016detecting}
Tian, Z., Huang, W., He, T., He, P., Qiao, Y.: Detecting text in natural image
  with connectionist text proposal network. In: Proceedings of the 11th
  European Conference on Computer Vision. pp. 56--72. Springer (2016)

\bibitem{vinyals2015grammar}
Vinyals, O., Kaiser, {\L}., Koo, T., Petrov, S., Sutskever, I., Hinton, G.:
  Grammar as a foreign language. In: Advances in Neural Information Processing
  Systems. pp. 2773--2781 (2015)

\bibitem{vinyals2015show}
Vinyals, O., Toshev, A., Bengio, S., Erhan, D.: Show and tell: A neural image
  caption generator. In: Proceedings of the IEEE conference on computer vision
  and pattern recognition. pp. 3156--3164 (2015)

\bibitem{wojna2017attention}
Wojna, Z., Gorban, A., Lee, D.S., Murphy, K., Yu, Q., Li, Y., Ibarz, J.:
  Attention-based extraction of structured information from street view
  imagery. arXiv preprint arXiv:1704.03549  (2017)

\bibitem{ijcai2017-458}
Xiao~Yang, Dafang~He, Z.Z.D.K.C.L.G.: Learning to read irregular text with
  attention mechanisms. In: Proceedings of the Twenty-Sixth International Joint
  Conference on Artificial Intelligence, {IJCAI-17}. pp. 3280--3286 (2017).
  \doi{10.24963/ijcai.2017/458}, \url{https://doi.org/10.24963/ijcai.2017/458}

\bibitem{yan2015deep}
Yan, F., Mikolajczyk, K.: Deep correlation for matching images and text. In:
  Proceedings of the IEEE conference on computer vision and pattern
  recognition. pp. 3441--3450 (2015)

\bibitem{yu2015large}
Yu, Q., Szegedy, C., Stumpe, M.C., Yatziv, L., Shet, V., Ibarz, J., Arnoud, S.:
  Large scale business discovery from street level imagery. arXiv preprint
  arXiv:1512.05430  (2015)

\bibitem{UberText}
Zhang, Y., Gueguen, L., Zharkov, I., Zhang, P., Seifert, K., Kadlec, B.:
  Uber-text: A large-scale dataset for optical character recognition from
  street-level imagery. In: SUNw: Scene Understanding Workshop - CVPR 2017.
  Hawaii, U.S.A. (2017)

\bibitem{zhang2016multi}
Zhang, Z., Zhang, C., Shen, W., Yao, C., Liu, W., Bai, X.: Multi-oriented text
  detection with fully convolutional networks. In: Computer Vision and Pattern
  Recognition,CVPR. Proceedings of the IEEE Computer Society Conference on
  (June 2016)

\bibitem{Zhou_2017_CVPR}
Zhou, X., Yao, C., Wen, H., Wang, Y., Zhou, S., He, W., Liang, J.: East: An
  efficient and accurate scene text detector. In: The IEEE Conference on
  Computer Vision and Pattern Recognition (CVPR) (July 2017)

\end{thebibliography}
\end{document}